\documentclass[conference]{IEEEtran}
\IEEEoverridecommandlockouts
\usepackage{cite}
\usepackage{amsmath,amssymb,amsfonts}
\usepackage{hyperref}
\usepackage{algorithmic}
\usepackage{graphicx}
\usepackage{colortbl}
\usepackage{textcomp}
\usepackage{xcolor}
\usepackage{subcaption}
\usepackage{mwe}
\def\BibTeX{{\rm B\kern-.05em{\sc i\kern-.025em b}\kern-.08em
    T\kern-.1667em\lower.7ex\hbox{E}\kern-.125emX}}
\begin{document}

\title{Pink-Eggs Dataset V1: A Step Toward Invasive Species Management Using Deep Learning Embedded Solutions}

\author{
\IEEEauthorblockN{Di Xu\textsuperscript{*}}
\IEEEauthorblockA{xudidee@163.com}
\and
\IEEEauthorblockN{Yang Zhao}
\IEEEauthorblockA{zy1192278004@gmail.com}
\and
\IEEEauthorblockN{Xiang Hao}
\IEEEauthorblockA{haoxiangsnr@gmail.com}
\and
\IEEEauthorblockN{Xin Meng}
\IEEEauthorblockA{ismenglx@163.com}
}

\maketitle
\begin{abstract}
We introduce a novel dataset consisting of images depicting pink eggs that have been identified as Pomacea canaliculata eggs, accompanied by corresponding bounding box annotations. The purpose of this dataset is to aid researchers in the analysis of the spread of Pomacea canaliculata species by utilizing deep learning techniques, as well as supporting other investigative pursuits that require visual data pertaining to the eggs of Pomacea canaliculata. It is worth noting, however, that the identity of the eggs in question is not definitively established, as other species within the same taxonomic family have been observed to lay similar-looking eggs in regions of the Americas. Therefore, a crucial prerequisite to any decision regarding the elimination of these eggs would be to establish with certainty whether they are exclusively attributable to invasive Pomacea canaliculata or if other species are also involved. The dataset is available at \url{https://www.kaggle.com/datasets/deeshenzhen/pinkeggs}.
\end{abstract}


\section{Introduction}
Pomacea canaliculata, which were introduced to various countries in the last century, are known to harbor hazardous bacteria and parasites [1,2]. Owing to their rapid growth and potential to invade new habitats, Pomacea canaliculata poses a significant ecological risk. Traditional methods for treating Pomacea canaliculata include pesticide development [3], importing natural enemies [4,5], and manual eradication. Recently, there has been considerable research interest in deep learning [6,7,8], which has demonstrated superior performance compared to humans in various tasks [9,10,11,12,13]. One proposed approach for the management of Pomacea canaliculata involves the use of laser rays as a potentially more energy-efficient method [14]. Nevertheless, the efficacy of this method is contingent on the availability of sufficient data to train deep learning models [15]. Hence, this report presents a collection of images featuring pink eggs identified as those spawned by Pomacea canaliculata, captured from multiple locations in Shenzhen, China. It is worth noting that the attribution of all pink eggs to Pomacea canaliculata cannot be definitively ascertained due to the lack of DNA testing and the resemblance between the eggs of Pomacea canaliculata and those of other species, thereby precluding a conclusive determination. Nevertheless, the presence of Pomacea canaliculata, their distinctive appearance, and the substantial quantity of pink eggs depicted in the images offer compelling evidence to suggest that the majority of the images feature Pomacea canaliculata eggs. 

\begin{figure}
    \centering
    {\renewcommand{\arraystretch}{0}
    \begin{tabular}{c@{}c}
    \begin{subfigure}[b]{.5\columnwidth}
        \centering
        \includegraphics[width=\columnwidth]{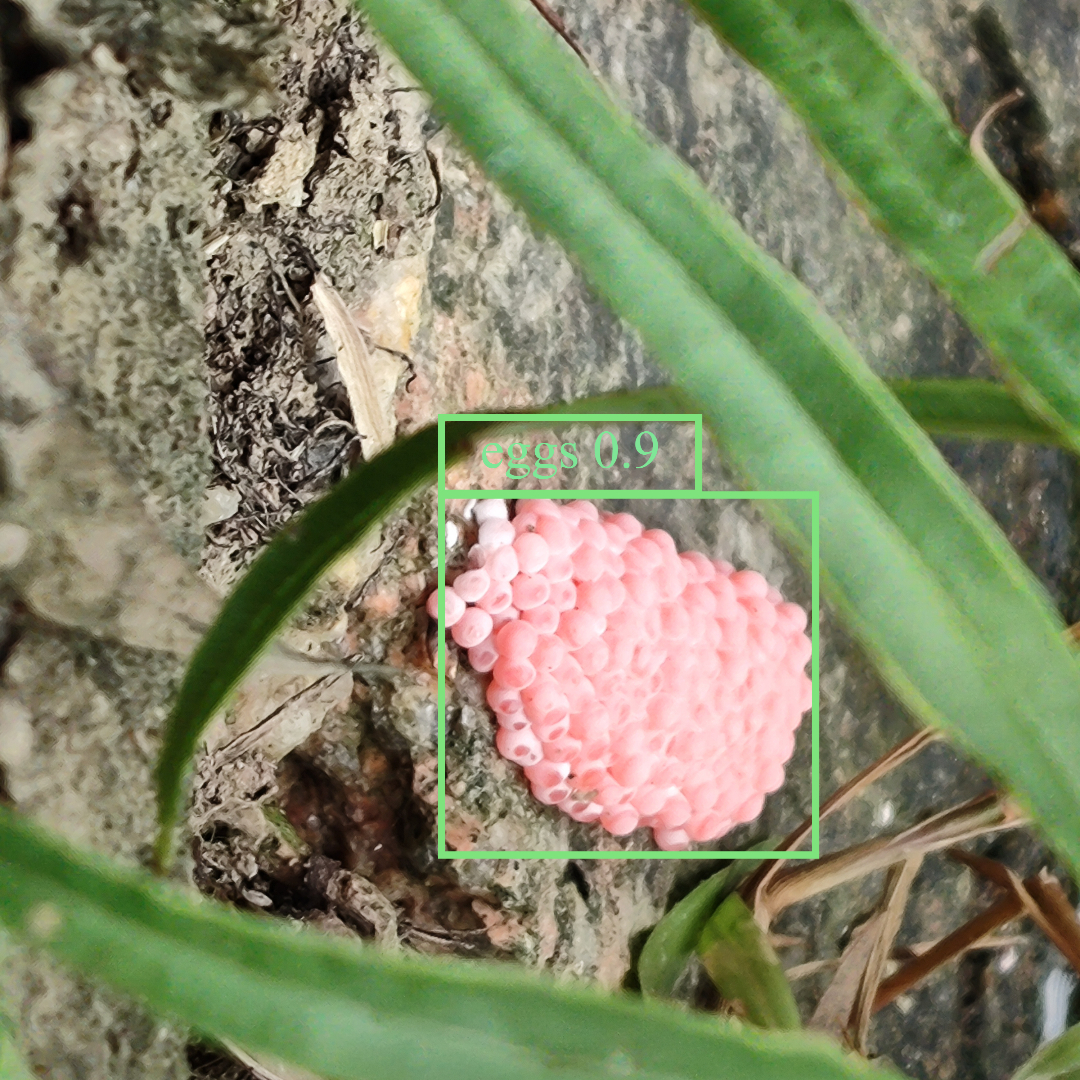}%
    \end{subfigure}&
    \begin{subfigure}[b]{.5\columnwidth}  
        \centering
        \includegraphics[width=\columnwidth]{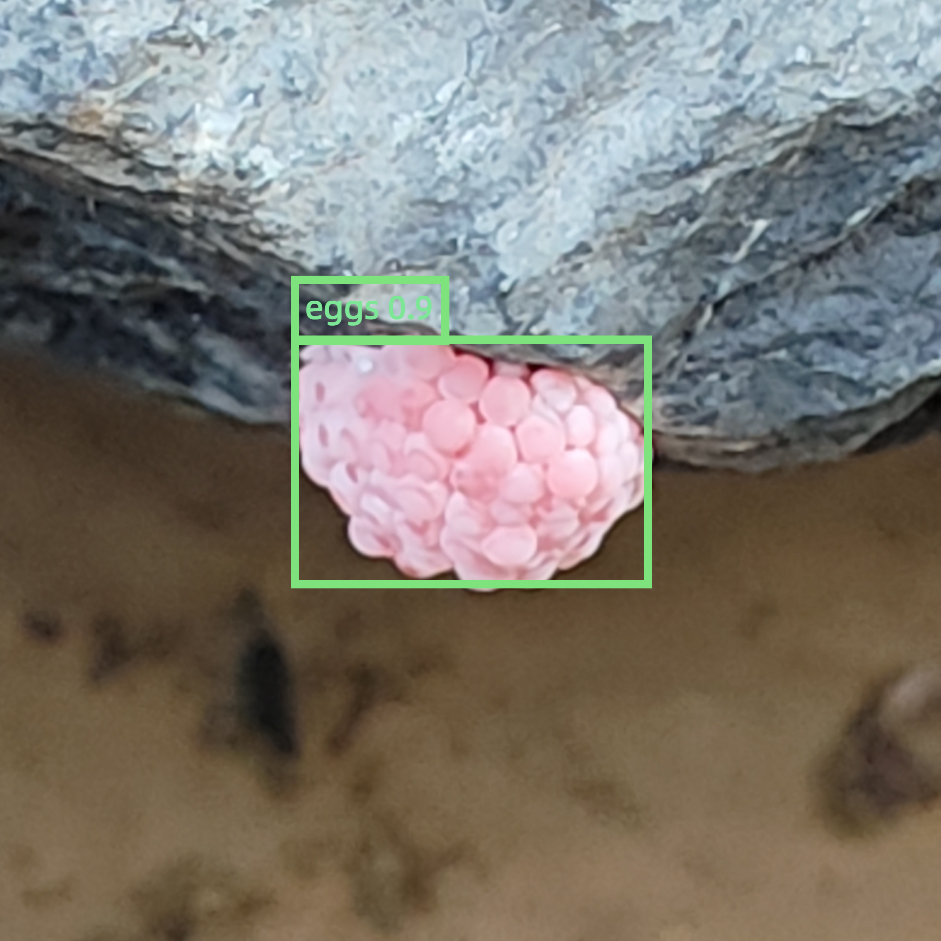}%
    \end{subfigure}\\
    \begin{subfigure}[t]{.5\columnwidth}   
        \centering 
        \includegraphics[width=\textwidth]{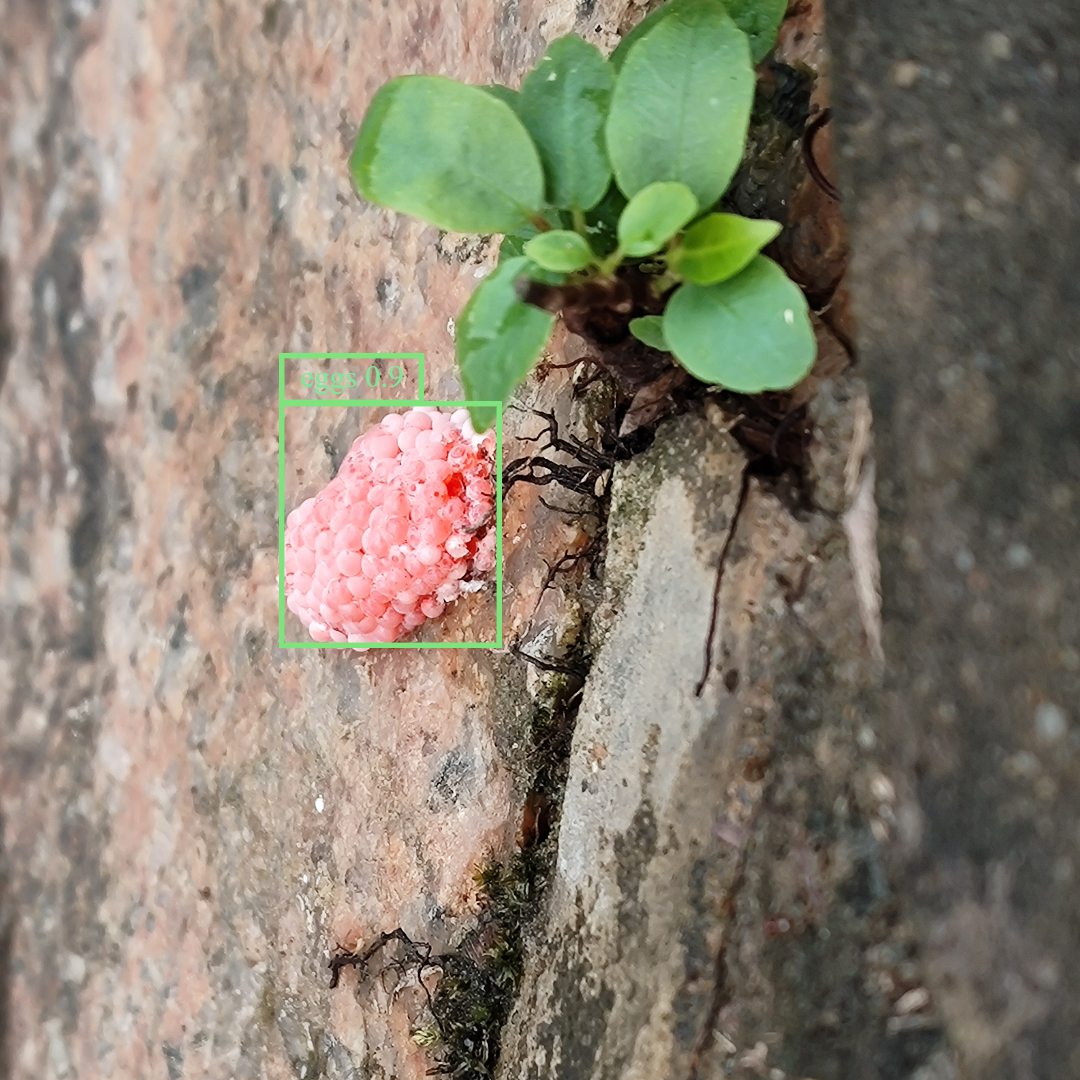}%
    \end{subfigure}&
    \begin{subfigure}[t]{.5\columnwidth}   
        \centering 
        \includegraphics[width=\columnwidth]{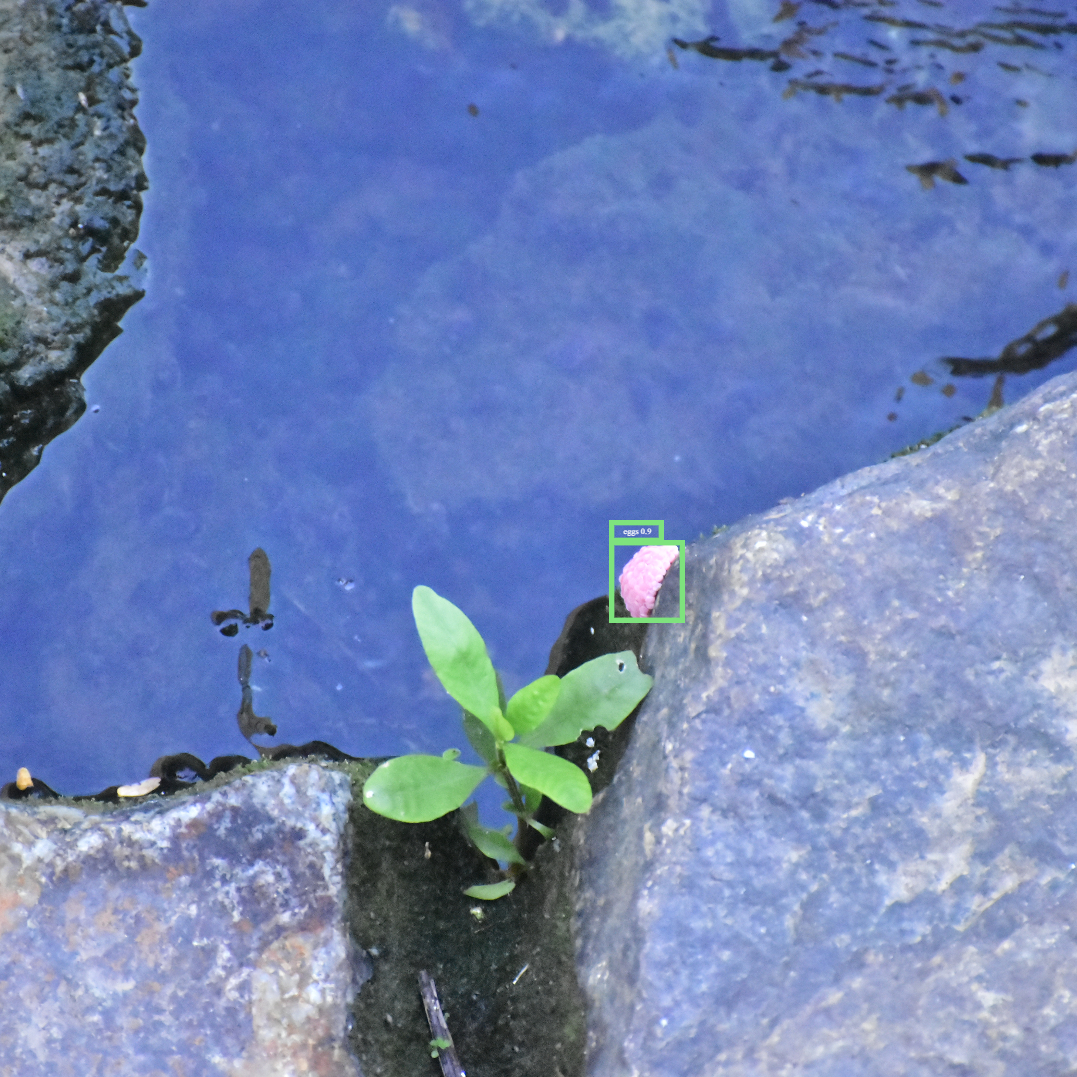}%
    \end{subfigure}
    \end{tabular}}
    \caption{Four examples of Pomacea canaliculata detection result with object bounding box localization.} 
\end{figure}

\section{Apple snails as invasive spices}

\subsection{Spreading of apple snails and the downsides}
Pomacea canaliculata, an apple snail species, is endemic to South America [1]. However, anthropogenic activities have facilitated its global dissemination during the past century. Owing to its prodigious reproductive rates and access to plentiful food resources, Pomacea canaliculata has flourished as an invasive species in the absence of natural predators, thereby posing a potential threat to indigenous species and disrupting wetland ecosystems. Furthermore, Pomacea canaliculata is known to act as a host for pathogenic bacteria and parasites, thereby presenting a risk to human health [2].

\subsection{Traditional ways to manage them}

Addressing the issue of invasive species, such as Pomacea canaliculata, can present significant challenges. Each approach to controlling or eradicating them entails a unique set of potential risks and benefits, which require careful consideration [3, 4, 5].

Among these approaches is the utilization of pesticides, which may exhibit efficacy in invasive species management but concurrently pose a considerable threat to non-target species within the surrounding environment [3]. Thus, a careful evaluation of the costs and benefits is imperative in determining the viability of pesticide application for invasive species management.

Artificial traps represent an additional approach for mitigating the population of Pomacea canaliculata [16, 17]. Nevertheless, the effectiveness of this method may be constrained in natural ecosystems lacking human regulation.

The introduction of natural predators as a means of controlling the population of Pomacea canaliculata is regarded as a feasible solution [4, 5]. However, it is imperative to conduct a comprehensive evaluation of the potential ecological impact before introducing non-native species. This is particularly crucial given that the introduction of predators for invasive species control has led to unintended consequences in some instances, as exemplified by the introduction of the cane toad in Australia [18].

\subsection{Targeting the eggs}

Pomacea canaliculata primarily inhabits underwater environments, and there is currently no evidence to suggest that they possess an innate tendency to conceal their eggs. However, these eggs are often deposited above the waterline, typically on hard surfaces such as rocks or vegetation in close proximity to the water's edge [19]. These eggs contain toxins and can be recognized by their distinct pink color, clustered arrangement, and variation in color intensity over time. Despite their lack of concealment, their conspicuous appearance facilitates easy detection by human observers. Consequently, the development of computer systems utilizing machine learning algorithms and computer vision techniques to recognize Pomacea canaliculata eggs based on their characteristic pink color and clustering pattern is a viable option. Such an approach would provide a more efficient and accurate means of identifying these eggs. 

A preliminary investigation was conducted to determine the origin of similar pink eggs that are found above the water line. The investigation targeted common freshwater snails and crustaceans in southern China, and it led to the conclusion that apple snails or their hybrids were the likely sources of the pink eggs, as depicted in Figure 1. While it cannot be definitely stated that all native species in Asia were examined, it is improbable that rare, unidentified, or undocumented species laid a significant number of eggs that resemble those of Pomacea canaliculata eggs. While it has been observed that other snail species in America deposit similar eggs [20], it is noteworthy that these species are also invasive in Asia. Therefore, additional research is imperative to authenticate the identification of the pink eggs detected near freshwater ponds prior to initiating any measures for their eradication. This inquiry could provide valuable information regarding the distribution and ecology of diverse snail species, thereby assisting in conservation and management endeavors. 


\subsection{Giving deep learning a try}
Deep neural networks, a type of machine learning method that employs the backpropagation algorithm, have demonstrated significant success in various tasks and have even surpassed human-level performance in certain instances [9,10,11,12,13]. Such tasks include computer vision applications such as image classification, object detection, and facial recognition, as well as tasks in diverse fields such as speech recognition, recommendation systems, and gaming. The noteworthy achievements of deep learning methods in diverse tasks offer promise for the development of advanced computer systems capable of completing complex tasks, such as object detection [21], object segmentation [22], and robot control [23]. 

Moreover, bespoke apparatus such as robotic vacuum cleaners or unmanned aerial vehicles [24] have demonstrated promise in the management of the invasive species Pomacea canaliculata. The amalgamation of technological advancements with environmental conservation initiatives has the potential to yield more effective and efficient solutions for addressing pressing ecological issues. Specifically, the emergence of embedded deep learning solutions for precision pesticide delivery has elicited substantial interest and holds potential for employment in the management of invasive species, including Pomacea canaliculata [25,26]. This approach has the potential to offer a more environmentally sustainable means of controlling invasive species by reducing the quantity of pesticide required for their management.

Besides the ecological benefits, the integration of embedded deep learning solutions has the potential to enhance management strategies by harnessing the power of data analytics and processing capabilities. These capabilities enable a comprehensive understanding of the behavior and population dynamics of invasive species such as Pomacea canaliculata, leading to more efficient resource utilization and minimizing negative environmental impacts.


Overall, embedded deep learning solutions hold great promise for the effective and sustainable management of invasive species, including Pomacea canaliculata. As such, further research in this area is warranted, with the potential to facilitate the creation of novel and innovative strategies for managing invasive species.

\section{Dataset}

\subsection{Details about the data}

Based on the morphological characteristics of the eggs observed, as well as the presence of Pomacea canaliculata in the surrounding area, we infer that the specimens captured in Shenzhen between October and December of 2022 during daylight hours and clear weather conditions are the eggs of Pomacea canaliculata.

For close-range photography, a Redmi K50 Ultra cellular device with default camera settings was employed to capture images of the eggs. To capture distant images, a D7200 camera equipped with an 18-140mm focus range lens was utilized in auto mode. In both cases, the images were saved in the JPG format.
     
After detecting distortions and imperfections in the collected data, data cleansing was performed by removing certain images that did not meet predetermined quality standards. Specifically, images that could be reliably identified as depicting Pomacea canaliculata eggs were retained, while images with severe degrees of blurriness were removed. These factors could be attributed to distance, motion-induced distortion, size, and camera-specific effects. 
               
Each image in the dataset was annotated with bounding box labels using the labelImg tool by three annotators, with all annotations incorporated into the dataset to facilitate object detection and classification. In order to minimize subjectivity in the labeling process, random samples of labeled data were reviewed. Furthermore, three sets of annotations were provided to enable evaluation through methods such as cross-validation and bootstrapping. The average Intersection over Union (IoU) rate was calculated between any two sets, and all the values are above 0.87. Based on these measures, we are highly confident that the collected images are suitable for supporting research hypotheses. 
               
The dataset was partitioned into three distinct subsets, namely train, dev, and test sets, each being mutually exclusive. The training set consisted of a total of 1000 images, randomly selected from the dataset, whereas the dev set and test set comprised 100 and 161 images, respectively. Additionally, all images were subjected to a modification process that involved the removal of embedded camera messages while preserving the original pixel values. 

In the pursuit of acquiring a comprehensive dataset, further images of Pomacea canaliculata eggs were sourced through online search engines. However, due to a dearth of explicit consent for their reuse, download, and distribution, these images could not be included in the final curation.


\subsection{Sample task}
The dataset under scrutiny harbors enormous potential for exploitation across a broad spectrum of domains, encompassing but not confined to image enhancement [27], super-resolution [28], object detection, and classification [21,22]. For instance, object detection is a convoluted process that entails recognizing and accentuating a target object within an image. In addition, the YOLOv5 model is a quintessential exemplar of a deep learning model that not only boasts unparalleled precision but also renders expeditious inference speed. Utilizing the dataset, this model has been judiciously trained to identify Pomacea canaliculata eggs, and its efficacy has been corroborated by demonstrative evidence [29,30]. Upon review of the demonstrated results, it can be deduced that the dataset contains a sufficient quantity of images paired with corresponding labels, capable of proficiently conducting Pomacea canaliculata eggs detection tasks even under adverse weather conditions. Nonetheless, it is imperative to note that the exhibited demonstration was not entirely optimized, warranting further experimentation to improve the model's precision and efficiency in identifying Pomacea canaliculata eggs.


\section{Conclusion}
By capitalizing on the Pomacea canaliculata eggs dataset, in tandem with the latest developments in deep learning methodologies, there is a promising opportunity to devise AI-embedded solutions to mitigate the impact of this invasive species. With the support of cutting-edge deep learning techniques, including object detection and classification models, it may yield precise and automated approaches to the detection and treatment of Pomacea canaliculata eggs in a myriad of environments. Furthermore, this approach is poised to diminish reliance on hazardous pesticides and arduous manual labor, thereby offering an environmentally sound and economically viable alternative. Therefore, fostering the development of AI-based solutions for invasive species management, particularly for Pomacea canaliculata, warrants research attention as it possesses the potential to yield favorable outcomes for our ecosystems.

\section*{Acknowledgment}

Thanks to Professor Robert Cowie and Jiwen Qiu for the responses on this topic.
Thanks to Hang Zhou, Qifan Feng and Xiaoqi Yu for their supports.


\end{document}